\documentclass[lettersize,journal]{IEEEtran}
\usepackage[strings]{underscore}
\usepackage{algorithmic}
\usepackage{algorithm}
\usepackage{amsmath}
\usepackage{array}
\usepackage{subcaption}
\usepackage{textcomp}
\usepackage{stfloats}
\usepackage{url}
\usepackage{verbatim}
\usepackage{graphicx}
\usepackage{booktabs}
\usepackage{cite}
\usepackage{subcaption}
\usepackage{float}
\usepackage{booktabs}
\usepackage{multirow}
\usepackage{tcolorbox}
\usepackage{amssymb}
\begin{document}

\title{Competent Autonomous Driving  Through Fine-Tuning Vision–Language–Action Models with Symbolic Traffic Rules}

\author{Barkin Dagda$^{1}$, Vinal Asodia$^{1}$, Iman Sharifi$^{2}$, Peng Wei$^{2}$, Saber Fallah$^{1}$
\thanks{$^{1}$Barkin Dagda, Vinal Asodia, and Saber Fallah are with the Connected and Autonomous Vehicles Lab (CAV-Lab), University of Surrey, Guildford, GU2 7XH, UK.%
        {\tt\small [bd00242, va00191, s.fallah]@surrey.ac.uk}}%
\thanks{$^{2}$Iman Sharifi and Peng Wei are with the Department of Mechanical and Aerospace Engineering, George Washington University, Washington, DC 20052, USA.%
        {\tt\small [i.sharifi, pwei]@gwu.edu}}%

}

\markboth{Journal of \LaTeX\ Class Files,~Vol.~14, No.~8, August~2021}%
{Shell \MakeLowercase{\textit{et al.}}: A Sample Article Using IEEEtran.cls for IEEE Journals}


\maketitle

\begin{abstract}
Vision-language-action (VLA) models offer a unified approach to perception, reasoning, and control for autonomous driving, but accurate action prediction alone does not establish compliance with traffic regulations. This paper presents a rule-guided VLA framework that incorporates symbolic traffic-rule feedback into policy optimisation. The framework combines supervised fine-tuning on expert demonstrations with reinforcement learning (RL) refinement using a reward that encourages encoded-rule compliance while penalising deviations from expert actions. Selected provisions of the United Kingdom Highway Code are formalised as executable first-order logic rules using PyReason, covering speed adaptation, following distance, and road-user priority. A frozen vision-language model extracts scene predicates, which are combined with vehicle-state information to assess the projected consequences of candidate control actions. Offline evaluation on held-out data collected in the CARLA simulator compares three group-based RL methods. The results show improvements in both action prediction accuracy and encoded-rule compliance over the supervised baseline. The best-performing variant increases the overall rule pass rate from 92.9\% to 97.6\%, with zero observed violations of the encoded road-user-priority check in the evaluated samples. These findings support symbolic rule feedback as a means of improving VLA policy alignment with explicit traffic-rule criteria.
\end{abstract}

\begin{IEEEkeywords}
Vision-language-action models, reinforcement learning, autonomous driving, first-order logic.
\end{IEEEkeywords}

\section{Introduction}

\IEEEPARstart{A}{}utonomous driving requires intelligent systems capable of perceiving complex environments, reasoning about dynamic traffic situations, and executing safe driving manoeuvres that comply with traffic regulations \cite{hu2026vla}. End-to-end learning addresses these interconnected tasks by mapping sensor observations to driving actions within a unified differentiable model. By jointly optimising perception and control, such approaches simplify system design and learn representations of complex driving behaviour \cite{chen2024e2e}. However, ensuring that learned policies comply with explicit traffic rules remains a challenge.

Large language models (LLMs) and vision-language models (VLMs) have expanded the role of semantic scene understanding and language-based reasoning in autonomous driving \cite{jiang2025,yildirim2025}. Frameworks such as DriveLM, LMDrive, and DriveGPT4 explore connections between scene understanding, reasoning, planning, and vehicle control \cite{drivelm,lmdrive,drivegpt4}. Vision-language-action (VLA) models further integrate multimodal observations, language, and action generation within a shared framework. Related models and benchmarks investigate the capabilities and limitations of these approaches in scene understanding, visual grounding, and driving decision-making \cite{guo2024drivemllm,jiang2024senna,xie2025drivebench,zhou2025opendrivevla,dvlm-ad,simlingo,cfvla,perceptwam}. Their development extends beyond imitation learning: AutoVLA employs reinforcement learning (RL) to refine planning and reasoning efficiency \cite{zhou2025autovla}, while AutoDrive-R$^2$ uses RL to improve reasoning and self-reflection \cite{autodrive}. Nevertheless, imitation objectives measure agreement with demonstrations, and RL objectives depend on the chosen rewards; neither inherently provides an explicit assessment of compliance with codified traffic rules.

Symbolic representations offer a means of making these requirements explicit. Traffic rules describe context-dependent constraints on driving behaviour, which can be encoded as logical conditions and evaluated against scene and vehicle-state information. Previous research has integrated symbolic knowledge into learning-based autonomous driving to support safety \cite{Sharifi_2}, interpretability \cite{Sharifi_1}, and policy alignment \cite{asodia2026}. SafeAuto \cite{safeauto}, for example, represents traffic regulations using Markov logic networks to assess whether generated behaviours satisfy specified constraints. These developments motivate connecting symbolic rule evaluation with VLA policy refinement. The challenge addressed here is to translate contextual traffic-rule checks into action-dependent feedback that directly guides continuous-control policy optimisation.

This paper proposes a rule-guided VLA framework that integrates symbolic traffic-rule evaluation into a two-stage learning procedure. Supervised fine-tuning (SFT) first adapts a pretrained VLM using expert driving actions and reference reasoning annotations. Reinforcement learning fine-tuning (RLFT) then refines the policy using a reward that combines rule-compliance feedback with a penalty for deviation from expert actions. Selected provisions of the United Kingdom (UK) Highway Code \cite{ukhighwaycode} are formalised as executable first-order logic (FOL) rules using the PyReason symbolic reasoning engine \cite{pyreason}. A separate frozen VLM extracts scene predicates, which are combined with vehicle-state information to evaluate the projected consequences of candidate actions. The resulting checks cover speed adaptation, following distance, and road-user priority, providing explicit learning signals for compliance with the encoded criteria.

We investigate this framework using three group-based optimisation methods: group relative policy optimisation (GRPO) \cite{deepseek,grpo}, group reward-decoupled normalisation policy optimisation (GDPO) \cite{gdpo}, and GRPO Done Right (Dr.~GRPO) \cite{drgrpo}. Offline experiments on held-out simulated driving data compare action prediction accuracy and encoded-rule compliance, while ablations examine the contribution of individual rules to policy refinement. The results show improvements over the SFT baseline in both evaluated objectives, supporting the use of symbolic rule feedback within the studied setting.

The main contributions are as follows:
\begin{itemize}
    \item We develop a VLA fine-tuning framework that incorporates symbolic traffic-rule feedback into policy optimisation while encouraging consistency with expert driving actions.

    \item We formalise selected UK Highway Code provisions as executable FOL rules and construct a symbolic evaluator that assesses candidate actions against context-dependent compliance criteria.

    \item We compare GRPO, GDPO, and Dr.~GRPO through offline evaluation and rule ablations, examining their effects on action prediction accuracy and compliance with the encoded rules.
\end{itemize}

Sections~\ref{sec:method} and~\ref{sec:symbolic} describe the learning framework and symbolic evaluator, respectively. Section~\ref{sec:experiments} presents the experimental setup, Section~\ref{sec:results} reports the results, and Section~\ref{sec:conclusion} concludes with limitations and future directions.
\begin{figure*}[t]
    \centering
    \includegraphics[width=\linewidth]{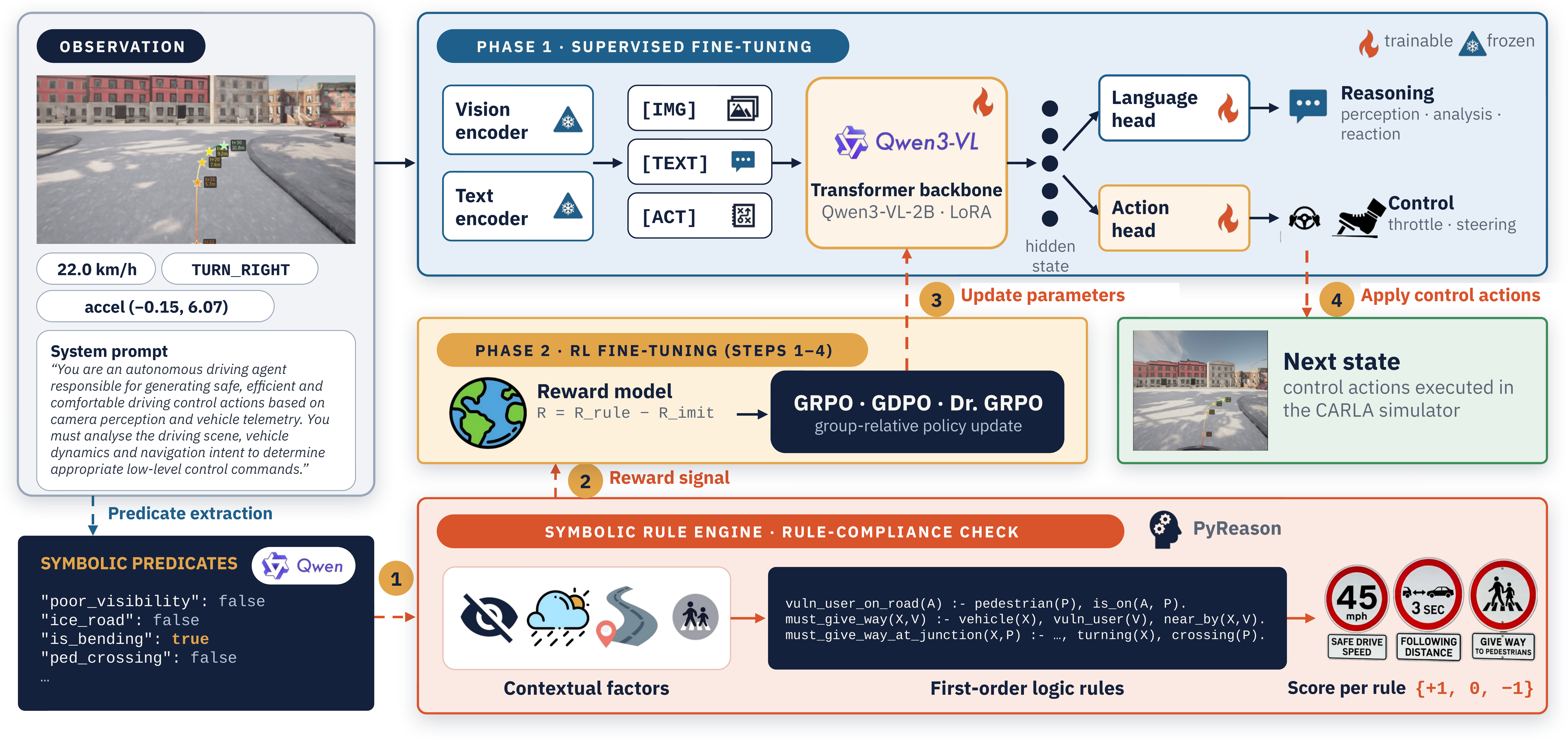}
    \vspace{-2mm}
    \caption{Overview of the proposed rule-guided VLA framework. SFT (top) trains the model to predict driving actions and generate reasoning outputs from visual observations and textual inputs. RLFT (bottom) refines the policy using symbolic traffic-rule feedback and an imitation penalty that encourages consistency with expert actions.}
    \label{fig:pipeline}
    \vspace{-3mm}
\end{figure*}
\section{Rule-Guided VLA Learning Framework}
\label{sec:method}

The proposed framework combines a VLA driving policy with a symbolic traffic-rule evaluator, as illustrated in Figure~1. Learning proceeds in two stages: SFT first adapts a pretrained VLM using expert actions and reference reasoning annotations, after which RL refines the policy using rule-compliance feedback and an imitation penalty. A separate frozen VLM extracts the scene predicates required for symbolic evaluation. 

\subsection{Problem Formulation}

At time $t$, the driving policy receives a visual observation $S_t$, the ego-vehicle state and telemetry $\mathbf{x}_t$, and a high-level navigation instruction $L$. The policy is represented by
\(\pi_\theta(\mathbf{a}_t,\mathbf{y}_t \mid S_t,\mathbf{x}_t,L)\),
where $\theta$ denotes its trainable parameters, $\mathbf{a}_t$ is a continuous vehicle-control action, and $\mathbf{y}_t$ is a natural-language reasoning output. The learning task is to produce driving actions that satisfy the applicable encoded traffic rules while remaining consistent with expert demonstrations.

Let $\mathcal{R}=\{r_1,r_2,\dots,r_K\}$ denote the set of encoded traffic rules. These rules depend on semantic properties of the driving scene, represented by the symbolic predicates $\mathbf{z}_t$ extracted from $S_t$ by the frozen VLM. Given these predicates, the vehicle-state information, and a candidate action, the symbolic evaluator determines which rules apply and whether the action satisfies them:
\( \mathbf{q}_t
=
Q(\mathbf{z}_t,\mathbf{x}_t,\mathbf{a}_t,\mathcal{R})
=
\left[q_{1,t},q_{2,t},\dots,q_{K,t}\right], \)
where $q_{k,t}$ records whether rule $r_k$ is inactive, satisfied, or violated under the supplied scene information and candidate-action evaluation. Only applicable rules contribute to the rule-compliance reward. The scene representation, action projection, and individual rule checks underlying $Q$ are specified in Section~\ref{sec:symbolic}.

The objective of rule-guided refinement is to improve compliance with these rules while penalising deviations from the corresponding expert actions. Accordingly, the policy-learning objective is formulated as
\begin{equation}
\theta^{*}
=
\arg\max_{\theta}
\mathbb{E}_{\pi_{\theta}}
\left[
R_t^{\mathrm{rule}}
-
\lambda R_t^{\mathrm{imit}}
\right],
\label{eq:policy_objective}
\end{equation}
where $R_t^{\mathrm{rule}}$ is derived from the compliance assessment $\mathbf{q}_t$, $R_t^{\mathrm{imit}}$ penalises deviation from the expert action $\mathbf{a}_t^{*}$, and $\lambda$ controls the relative weight of the imitation penalty. This objective allows explicit rule feedback to influence policy updates while encouraging consistency with demonstrated driving behaviour. 

\subsection{VLA Policy Architecture}
\label{sec:architecture}

The driving policy is built on the Qwen3-VL-2B-Instruct backbone. The visual observation $S_t$ is provided as an image, while the ego-vehicle state and telemetry $\mathbf{x}_t$ are converted into textual descriptions and combined with the navigation instruction $L$ to form the accompanying prompt. These inputs are encoded as visual and text tokens within a shared multimodal sequence. A learnable action token, denoted as \texttt{[ACT\_TOKEN]}, is added to the vocabulary and appended to the input sequence to provide a dedicated representation for continuous action prediction.

The multimodal sequence is processed by the transformer backbone, which integrates the visual, vehicle-state, and navigation information through self-attention. The resulting contextualised hidden states support both action prediction and language generation. In particular, the hidden state associated with \texttt{[ACT\_TOKEN]} represents the driving context used by the action head.

Two output heads generate the vehicle-control action and the natural-language reasoning output. The action head is a multilayer perceptron with Gaussian error linear unit (GELU) activations that maps the action-token hidden state to
\( \mathbf{a}_t
=
(a_{t,\mathrm{throttle}},a_{t,\mathrm{steering}})
\in [0,1]\times[-1,1]. \)
A sigmoid activation bounds the throttle output to $[0,1]$, while a $\tanh$ activation bounds the steering output to $[-1,1]$. The language-modeling head autoregressively generates the reasoning output $\mathbf{y}_t$ from the corresponding language-token hidden states. These two outputs are supervised using expert control actions and reference reasoning annotations, respectively, as described next.

\vspace{-3mm}
\subsection{Supervised Fine-Tuning}
Supervised fine-tuning adapts the policy using driving observations paired with expert control actions and reference reasoning annotations. The demonstration dataset is denoted by
\( \mathcal{D}_{\mathrm{expert}}
=
\left\{
(S_t,\mathbf{x}_t,L,\mathbf{a}_t^{*},\mathbf{c}_t^{*})
\right\}_{t=1}^{T}, \)
where $T$ is the number of training samples, $\mathbf{a}_t^{*}$ is the expert action, and $\mathbf{c}_t^{*}$ is the reference reasoning annotation associated with the observation. These targets jointly supervise the policy's action and language outputs, establishing an initial behavioural and reasoning prior.

The action loss measures the squared error between the predicted control vector $\hat{\mathbf{a}}_t$ and the expert action:
\begin{equation}
\mathcal{L}_{\mathrm{action}}(\theta)
=
\frac{1}{T}
\sum_{t=1}^{T}
\left\|
\hat{\mathbf{a}}_t-\mathbf{a}_t^{*}
\right\|_2^2.
\label{eq:action_loss}
\end{equation}
The language loss supervises autoregressive generation of the reference reasoning annotation:
\begin{equation}
\mathcal{L}_{\mathrm{reason}}(\theta)
=
-\frac{1}{T}
\sum_{t=1}^{T}
\sum_{k=1}^{K_t}
\log p_\theta
\left(
c_{t,k}^{*}
\mid
S_t,\mathbf{x}_t,L,c_{t,<k}^{*}
\right),
\label{eq:reason_loss}
\end{equation}
where $K_t$ is the annotation length, $c_{t,k}^{*}$ is its $k$-th token, and $c_{t,<k}^{*}$ denotes the preceding tokens. The two losses are combined as
\( \mathcal{L}_{\mathrm{SFT}}(\theta)
=
\mathcal{L}_{\mathrm{action}}(\theta)
+
\mathcal{L}_{\mathrm{reason}}(\theta). \)
This objective encourages the policy to reproduce expert controls and generate reasoning outputs consistent with the reference annotations. The resulting SFT checkpoint provides the initialisation for the rule-guided reinforcement-learning stage.

\vspace{-3mm}
\subsection{Rule-Guided Reinforcement Learning}
\label{sec:optimization}

Following SFT, the policy is refined using symbolic rule feedback together with an imitation penalty. For each driving observation, the deterministic action prediction is treated as the mean of a Gaussian distribution,
$\mathcal{N}(\boldsymbol{\mu}_{\theta}(S_t,\mathbf{x}_t,L),\Sigma)$,
where $\Sigma$ is the action-sampling covariance. A group of $G$ candidate actions
$\{\mathbf{a}_{t,i}\}_{i=1}^{G}$
is sampled from this distribution. Evaluating alternatives within the same driving context provides the relative feedback used to update the policy.

Each candidate action is projected over a short horizon and assessed by the symbolic evaluator using the resulting vehicle-state information and the scene predicates $\mathbf{z}_t$. The projection procedure and individual rule checks are detailed in Section~\ref{sec:symbolic}. For each rule $r\in\mathcal{R}$, the evaluator assigns
\begin{equation}
s_r(\mathbf{a}_{t,i})
=
\begin{cases}
-1, & \text{rule violated}, \\
0,  & \text{rule inactive}, \\
+1, & \text{rule satisfied}.
\end{cases}
\label{eq:rule_score_case}
\end{equation}
Let $\mathcal{R}_{\mathrm{active}}\subseteq\mathcal{R}$ denote the rules applicable to the current scene, with
$n_{\mathrm{active}}=|\mathcal{R}_{\mathrm{active}}|$.
The rule-compliance reward averages the scores over these active rules and includes a small bonus proportional to the fraction satisfied:
\begin{equation}
R_{t,i}^{\mathrm{rule}}
=
\frac{1}{n_{\mathrm{active}}}
\sum_{r\in\mathcal{R}_{\mathrm{active}}}
s_r(\mathbf{a}_{t,i})
+
\beta
\frac{n_{\mathrm{pass},t,i}}{n_{\mathrm{active}}},
\label{eq:rule_reward}
\end{equation}
where $n_{\mathrm{pass},t,i}$ is the number of active rules satisfied by candidate $i$ and $\beta=0.05$. The bonus slightly increases the contribution of rule satisfaction to the reward.

To retain consistency with demonstrated driving behaviour, an imitation penalty measures the candidate's deviation from the corresponding expert action:
\begin{equation}
R_{t,i}^{\mathrm{imit}}
=
\left\|
\mathbf{a}_{t,i}-\mathbf{a}_t^{*}
\right\|_2^2.
\label{eq:imit_reward}
\end{equation}
The two terms form the composite reward
\begin{equation}
R_{t,i}
=
R_{t,i}^{\mathrm{rule}}
-
R_{t,i}^{\mathrm{imit}},
\label{eq:composite_reward}
\end{equation}
which corresponds to the unit-weight case, $\lambda=1$, of Equation~\ref{eq:policy_objective}. This reward encourages satisfaction of the encoded rules while penalising deviations from expert controls. Mission completion, trajectory smoothness, and ride comfort are not explicitly included in the reward; the imitation term encourages consistency with demonstrations without directly constraining these temporal properties.

Policy updates use a clipped, group-relative surrogate objective that requires no separate critic network. For a candidate sampled from the previous policy $\pi_{\theta_{\mathrm{old}}}$, the likelihood ratio is
\begin{equation}
\rho_{t,i}
=
\frac{
\pi_{\theta}(\mathbf{a}_{t,i}\mid S_t,\mathbf{x}_t,L)
}{
\pi_{\theta_{\mathrm{old}}}(\mathbf{a}_{t,i}\mid S_t,\mathbf{x}_t,L)
}.
\end{equation}
The shared clipped surrogate is
\begin{equation}
\begin{aligned}
\mathcal{J}_{\mathrm{clip}}(\theta)
=
\frac{1}{G}
\sum_{i=1}^{G}
& \min
(
\rho_{t,i}A_{t,i}, \\
&\operatorname{clip}
\left(
\rho_{t,i},
1-\epsilon_{\mathrm{clip}},
1+\epsilon_{\mathrm{clip}}
\right)
A_{t,i}
),
\label{eq:grpo_general}
\end{aligned}
\end{equation}
where $A_{t,i}$ is the candidate's group-relative advantage and $\epsilon_{\mathrm{clip}}$ controls the clipping interval. GRPO and GDPO additionally use KL regularisation relative to the SFT policy, while this term is omitted for Dr.~GRPO.

The three variants differ in how they construct advantages from the candidate rewards. Let
$\mathbf{R}_t=(R_{t,1},\dots,R_{t,G})$
denote the rewards for a sampled group:
\begin{itemize}
    \item \textbf{GRPO} centres the composite rewards and normalises them by their within-group standard deviation:
    \[
    A_{t,i}
    =
    \frac{
    R_{t,i}-\operatorname{mean}(\mathbf{R}_t)
    }{
    \operatorname{std}(\mathbf{R}_t)+\epsilon
    },
    \]
    where $\epsilon$ is a small constant for numerical stability.

    \item \textbf{GDPO} centres and normalises the rule-compliance and imitation components separately before combining their advantages:
    \( A_{t,i}
    =
    A_{t,i}^{\mathrm{rule}}
    -
    A_{t,i}^{\mathrm{imit}}. \)
    This separates the normalisation of the two components so that their different scales do not directly determine their relative contributions.

    \item \textbf{Dr.~GRPO} centres the composite rewards without dividing by their standard deviation:
    \( A_{t,i}
    =
    R_{t,i}
    -
    \operatorname{mean}(\mathbf{R}_t). \)
\end{itemize}
These updates favour candidate actions according to their relative rule compliance and imitation error. 

\section{Symbolic Traffic-Rule Representation and Evaluation}
\label{sec:symbolic}

The symbolic evaluator provides the rule-compliance feedback used by the learning framework described in Section~\ref{sec:method}. It operationalises selected UK Highway Code provisions covering speed adaptation, following distance, and road-user priority. Evaluation proceeds by extracting symbolic scene information, inferring contextual factors, and applying the encoded rules to the projected consequences of candidate actions. The following subsections describe these components and how they produce the compliance assessments used during training and evaluation.

\vspace{-3mm}
\subsection{Symbolic Scene Representation}
The information required for symbolic evaluation is identified through manual analysis of the selected Highway Code provisions. This analysis defines a predicate vocabulary describing road geometry, environmental conditions, nearby road users, and vehicle interactions. Qualitative scene properties are represented by these predicates, while numerical quantities required for action evaluation, such as ego speed and following distance, are supplied separately through the vehicle-state information $\mathbf{x}_t$.

A frozen Qwen2.5-VL-7B-Instruct model extracts the predicates from the visual observation:
\( \mathbf{z}_t = f_{\phi}(S_t), \)
where $f_{\phi}$ is the fixed visual-to-symbolic mapping and $\mathbf{z}_t$ contains the rule-relevant predicates for the observed scene. Structured prompts specify the meanings of the predicates and consistency requirements, such as mutually exclusive road-surface labels. The model's output is parsed into a dictionary of Boolean predicates that serves as input to the symbolic reasoning engine.

\begin{figure}[t]
    \centering
    \includegraphics[width=\linewidth, trim={12mm 12mm 12mm, 12mm}, clip]{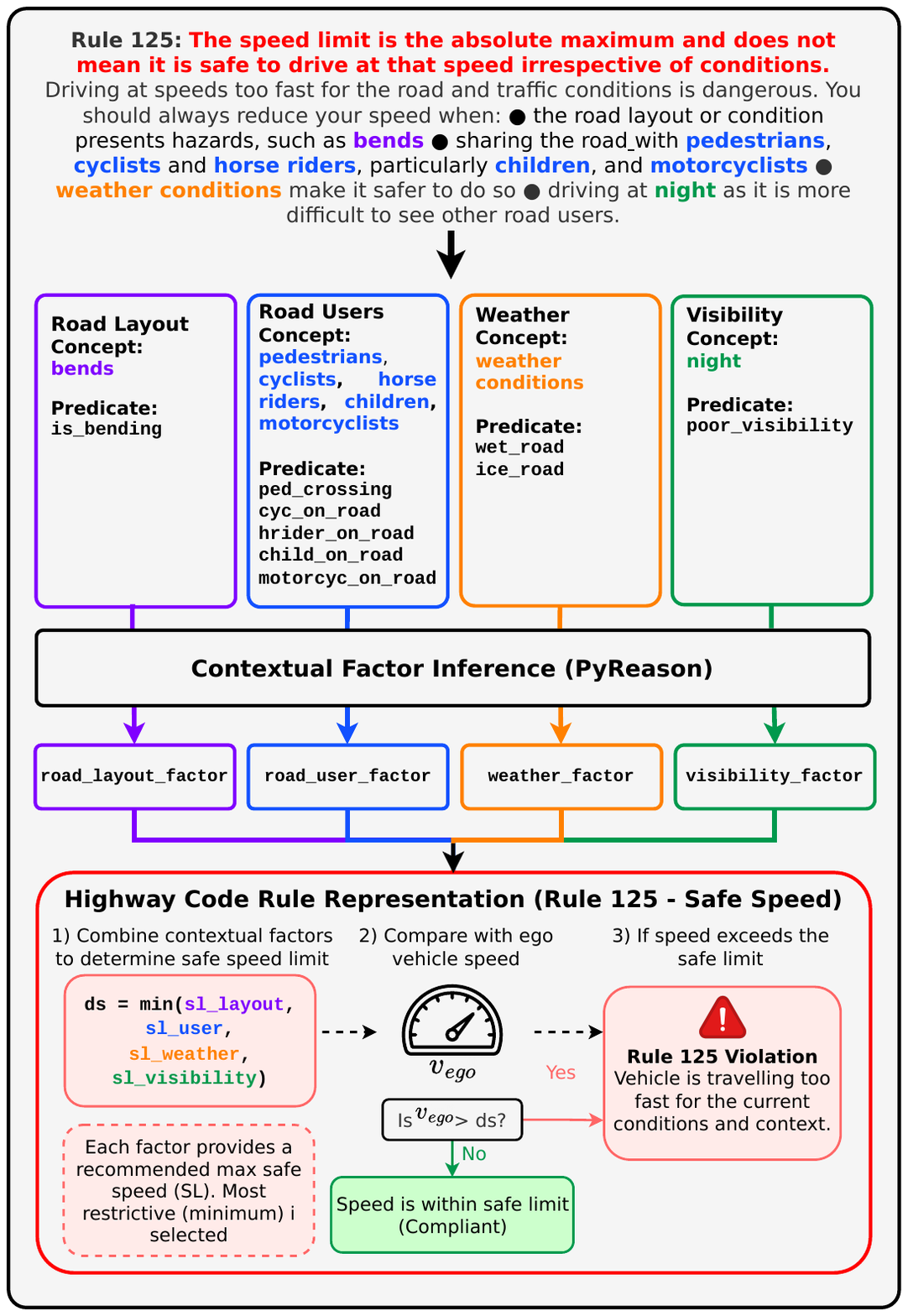}
    \caption{An illustrative example of translating a UK Highway Code rule into symbolic predicates, contextual factors, and an executable compliance check.}
    \label{fig:rule_creation}
    \vspace{-3mm}
\end{figure}

PyReason applies logical inference rules to these predicates to derive the contextual factors used by the compliance checks. This step connects individual scene attributes to the parameters of the encoded traffic rules. For example, predicates indicating a wet road or poor visibility inform the factors used to adjust speed and following-distance thresholds. Figure~\ref{fig:rule_creation} illustrates the progression from traffic-rule concepts to predicates, contextual factors, and an executable check. The resulting factors parameterise the rule formalizations described next.
\subsection{Formalization of Selected Traffic Rules}
The contextual factors derived from the scene representation parameterise three compliance checks: speed adaptation, safe following distance, and road-user priority. These checks operationalise selected Highway Code provisions through explicit conditions and thresholds. The equations below define the framework's numerical interpretation of those provisions.

\noindent
\textbf{Rule 125 -- Safe Speed:}
Rule 125 in UK Highway Code advises drivers to adapt their speed to the prevailing road, traffic, and environmental conditions, while Rule 124 specifies the applicable maximum speed limits. The contextual speed check uses factors associated with road layout, nearby road users, weather, and visibility to determine a desired speed:
\( DS = \min\bigl(
SL_{\mathrm{layout}}, SL_{\mathrm{users}},
SL_{\mathrm{weather}}, SL_{\mathrm{visibility}}
\bigr), \)
where each $SL$ term denotes the speed threshold associated with the corresponding contextual factor. The minimum selects the most restrictive of these thresholds. A violation is recorded when the ego speed evaluated for a candidate action exceeds $DS$.

\noindent
\textbf{Rule 126 -- Safe Following Distance:}
Rule 126 in UK Highway Code addresses the separation needed to stop safely if a preceding vehicle slows or stops, accounting for the driving conditions. Within the framework, a base stopping distance is adjusted using factors associated with traffic, visibility, road conditions, and vehicle characteristics:
\( SD = BSD \times \max\bigl(
F_{\mathrm{traffic}}, F_{\mathrm{visibility}},
F_{\mathrm{road}}, F_{\mathrm{vehicle}}
\bigr), \)
where $BSD$ is the base stopping distance under normal conditions and the $F$ terms are the contextual scaling factors. This aggregation applies the largest individual multiplier to obtain the required distance $SD$. The check records a violation when the following distance evaluated for the candidate action is less than $SD$.

\noindent
\textbf{Rule H -- Hierarchy of Road Users:}
According to UK Highway Code, the hierarchy principle in Rule H1 assigns greater responsibility for reducing danger to road users capable of causing greater harm, while Rule H2 specifies priority guidance for pedestrians and other vulnerable road users. The framework implements a selected set of yielding conditions using information about junctions, crossings, road users, and traffic signals:
\begin{itemize}
    \item pedestrians crossing or waiting to cross the road into or out of which the ego vehicle is turning;
    \item pedestrians at zebra crossings, and pedestrians or cyclists at parallel crossings;
    \item pedestrians at light-controlled crossings when the signal controlling the ego vehicle is red.
\end{itemize}
These conditions define the scope of the implemented road-user-priority check.

For each situation, the symbolic predicates describing road-user location, crossing intent, and the relevant road or signal context determine whether a yielding response is required. Vehicle-motion information is then used to assess whether the candidate action initiates the required deceleration. A violation is recorded when the yielding condition is active but this response is absent. This separates the applicability of the check from the assessment of the vehicle's response. The following subsection explains how the evaluator applies these checks to the projected consequences of candidate actions.

\vspace{-3mm}
\subsection{Candidate-Action Evaluation}
To assess the consequences of a candidate action without executing it in the simulator during fine-tuning, the evaluator projects the ego-vehicle state over a short horizon $\Delta t$ using a kinematic bicycle model~\cite{rajamani2012vehicle}. For candidate $\mathbf{a}_{t,i}$, this projection is represented as
\( \hat{\mathbf{x}}_{t+1,i}
=
\mathcal{F}_{\Delta t}
\left(
\mathbf{x}_t,\mathbf{a}_{t,i}
\right), \)
where $\mathcal{F}_{\Delta t}$ denotes the state-projection model and $\hat{\mathbf{x}}_{t+1,i}$ is the predicted ego-vehicle state at the end of the horizon. The model provides a computationally inexpensive approximation of vehicle motion, while omitting higher-order dynamics such as tyre slip.

The projected state is evaluated together with the symbolic predicates $\mathbf{z}_t$ and the relevant vehicle-state information. Candidates sampled for the same observation share the extracted scene predicates, while their projected states reflect their different control actions. PyReason uses the scene predicates and inferred contextual factors to identify the applicable rules and their parameters. The candidate's projected motion is then assessed against the corresponding criteria defined in the preceding subsection. Rules whose activation conditions are absent remain inactive and are excluded from the rule-compliance reward.

For each candidate, this procedure produces a compliance vector $\mathbf{q}_{t,i}$ through the evaluator $Q$ introduced in Section~\ref{sec:method}. Its entries are mapped to the satisfied, violated, or inactive scores defined in Equation~\ref{eq:rule_score_case}, which supply the rule-compliance reward in Equation~\ref{eq:rule_reward}. The same checks support the offline assessment of predicted actions. Their outputs indicate satisfaction of the encoded criteria under the supplied scene information and the state-projection assumptions. The following section describes the data collection, training configuration, and evaluation protocol used to study the framework.

\begin{figure*}[h]
    \centering
    \includegraphics[scale=0.50]{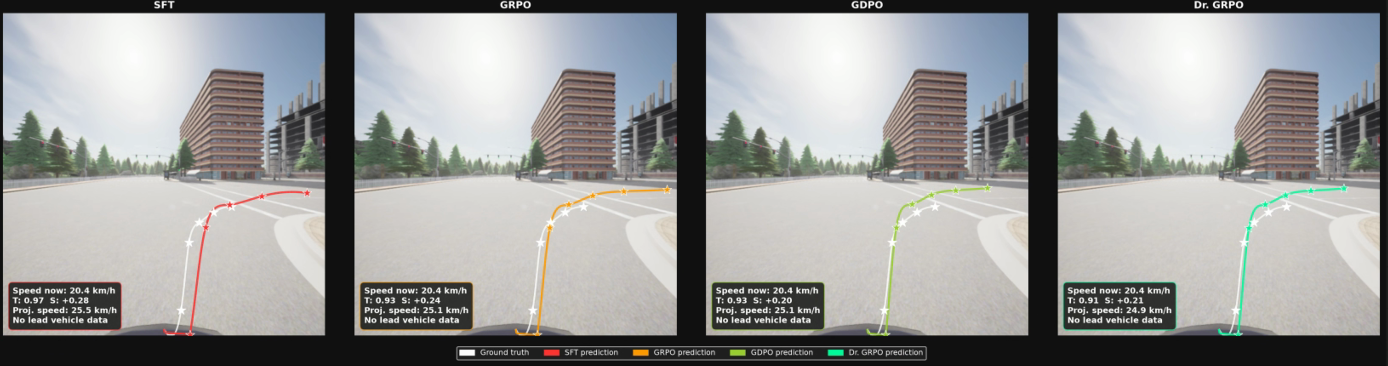}
    \caption{An illustrative example of the trajectory predictions made by the SFT, GRPO, GDPO and Dr. GRPO pretrained VLA models. The figures show the throttle, steering, projected speed and the trajectory waypoints. }
    \label{fig:traje_comparison}
\end{figure*}

\section{EXPERIMENTAL SETUP} \label{sec:experiments}
\subsection{Data Collection}
\noindent
To train and evaluate a VLA model with traffic-rule compliance, scene attributes serve as key inputs for the symbolic rule engine. The CARLA simulation environment \cite{carla} provides ground-truth scene attributes, including road conditions, road geometry, the presence of nearby vehicles, and the proximity of vulnerable road users. To collect the expert driving data, an automated rule-based agent navigates across multiple CARLA towns, covering residential, urban, highway, and multi-lane environments under clear and rainy weather conditions, along with dedicated pedestrian-crossing scenarios.
The agent operates with a two-level planning architecture: at the global level, it calculates optimal routes across the road network to generate target waypoints and high-level navigation commands (e.g., \textit{LANEFOLLOW}, \textit{TURNLEFT}, \textit{TURNRIGHT}). At the local level, it tracks these waypoints and manages speed using proportional-integral-derivative controllers \cite{astrom2021feedback} while reacting to red lights and leading obstacles. At each time step, the system records synchronized data, including a front-facing RGB image, vehicle kinematic states (speed, heading, and acceleration), distance to leading vehicles, the high-level navigation commands, the executed steering, throttle, and braking actions.

To extract scene attributes directly from raw perceptual data for the symbolic rule engine, each collected front-facing RGB image is processed using a frozen pre-trained Qwen2.5-VL-7B-Instruct model. The model is guided by structured prompt instructions with explicit semantic definitions and logical consistency constraints, such as ensuring road surface conditions remain mutually exclusive and asserting pedestrian presence whenever crossing behavior is identified. Each visual frame is parsed into a verified dictionary of boolean predicates representing environmental visibility, road geometry, obstacle categories, and vulnerable road users.

Reasoning annotations ($c_t^*$) are created automatically by combining vehicle sensor data with information about the scene. For each frame, the system combines measurements such as the vehicle speed compared with the speed limit, IMU acceleration, the curvature of the upcoming road, traffic light signals, and the distance to pedestrians with scene information extracted by the vision-language model.

These inputs are organized into three reasoning steps:

\begin{itemize}
    \item \textbf{Perception:} Describes what is happening around the vehicle. It includes the road condition, weather and visibility, traffic light state, detected obstacles, and the distance to pedestrians or other vulnerable road users.

    \item \textbf{Analysis:} Explains the driving situation by considering the planned route and safety conditions. It checks the road curvature using the projected waypoints, compares the vehicle speed with the speed limit, and identifies important safety situations, such as a red traffic light or a nearby pedestrian.

    \item \textbf{Reaction:} Explains what the vehicle should do next. It describes whether the vehicle should accelerate, maintain its motion, coast, or brake. It also explains any steering changes needed to safely follow the planned path.
\end{itemize}

To avoid having too many similar frames when the vehicle is stopped for a long time, a temporal filter removes repeated stationary frames. Important frames, such as the start of braking and the beginning of movement, are kept.

Finally, the generated reasoning text is combined with the vehicle-state prompt, the input image, and the continuous driving actions. Together, these form the final multi-modal training samples.

\vspace{-3mm}
\subsection{Training \& Testing}
The Qwen3-VL-2B-Instruct backbone is fine-tuned using LoRA adapters applied to the attention projection layers (\texttt{q\_proj}, \texttt{k\_proj}, \texttt{v\_proj}, and \texttt{o\_proj}), with a rank of $16$, a scaling factor of $32$, and a dropout rate of $0.05$.
SFT is conducted on a dataset of approximately 20,000 CARLA images partitioned into an 80/20 train-validation split. Training is performed over four epochs with a batch size of $16$ and a maximum sequence length of $2048$ tokens. Optimization is governed by the AdamW optimizer with a learning rate of $2 \times 10^{-5}$, utilizing a cosine learning-rate scheduler, gradient clipping, mixed-precision training, and gradient check-pointing to maximize memory efficiency.
To jointly optimize the VLA model for both continuous vehicle control and interpretable driving reasoning, a multi-task objective is formulated, consisting of an action prediction loss and a language reasoning loss. The action prediction loss is defined via MSE, which minimizes the discrepancy between the predicted control commands and the expert trajectories. Concurrently, the language reasoning loss trains the model to generate scene-grounded explanations through auto-regressive next-token prediction over the ground-truth captions.
RLFT was initialized from the pretrained SFT checkpoint and optimized using AdamW with a learning rate of $1 \times 10^{-6}$. For each training sample, four candidate action trajectories were generated to estimate relative advantages, and KL regularization was applied to constrain policy drift from the supervised initialization in GRPO and GDPO. RL training was conducted for 15 epochs, a duration determined empirically by monitoring reward curves. All experiments are conducted on a machine equipped with NVIDIA RTX 5070 Ti GPU, with 16 GB VRAM, using the same effective batch size and mixed-precision configuration as SFT.
The SFT-pretrained and RLFT VLA policies are evaluated with approximately 3,600 image test dataset. The test dataset includes selected driving scenarios to ensure that each of the three active Highway Code rules is triggered across a sufficient number of frames for statistically meaningful per-rule compliance estimation. The test set covers a diverse range of driving scenarios, including pedestrian crossings, rainy weather conditions, and highway driving. For each scenario, the VLA model is tasked with predicting vehicle actions over the next five time steps, thereby generating a future vehicle trajectory. Figure~\ref{fig:traje_comparison} presents an illustrative comparison of the trajectories predicted by the SFT-pretrained and RLFT VLA policies, together with the estimated vehicle speed and the predicted throttle and steering commands.

\vspace{-3mm}
\subsection{Evaluation Metrics}
Four complementary evaluation criteria are used to assess the performance of the SFT and RLFT policies. Control accuracy is measured via mean absolute error (MAE) and root mean square error (RMSE) for throttle and steering commands, quantifying how closely the predicted actions match expert demonstrations. Aggregate rule compliance is reported as an overall pass rate and violation rate across all test samples, providing a measure of how consistently the policy satisfies the active UK Highway Code rules. Per-rule compliance is assessed using the PyReason rule score defined in Equation~\ref{eq:rule_score_case}, which assigns a graded value in $[-1, +1]$ for each rule averaged only over frames in which that rule is active. Scenario-specific compliance is further examined through a condition-level breakdown, isolating policy behaviour under individual driving scenarios that activate each rule's predicates. Finally, an ablation study is conducted using the same evaluation protocol, with each UK Highway Code rule systematically removed from the composite reward signal, to quantify the individual contribution of each rule to the learned policy's compliance behaviour.

\begin{figure}[t]
    \centering
    \begin{subfigure}[b]{\linewidth}
        \centering
        \includegraphics[width=1.0\linewidth]{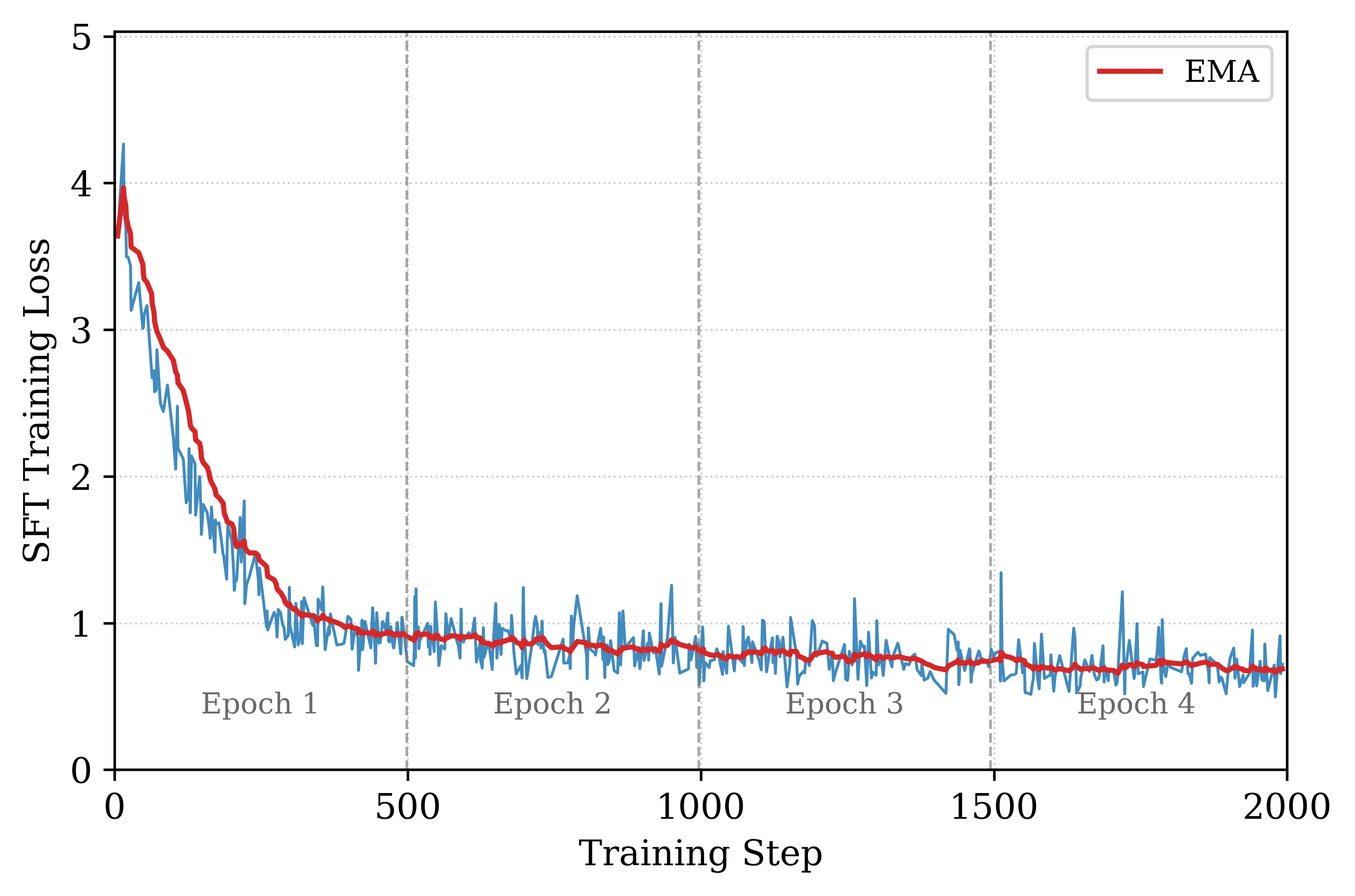}
        \caption{Exponential Moving Average (EMA) of SFT training loss.}
        \label{fig:sft_loss_curve}
    \end{subfigure}

    \begin{subfigure}[b]{\linewidth}
        \centering
        \includegraphics[width=1.0\linewidth]{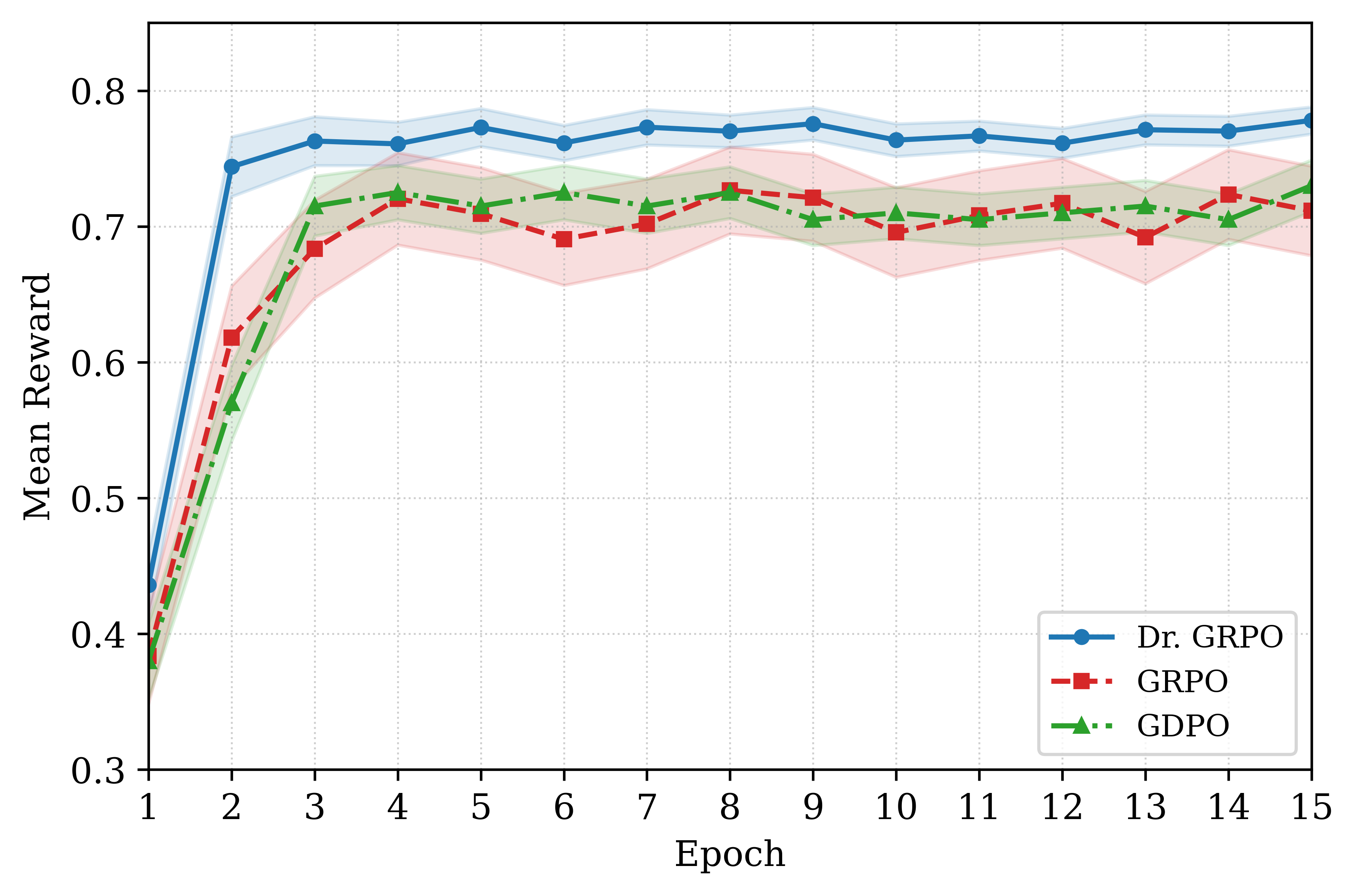}
        \caption{Mean reward over 15 epochs for RL fine-tuning.}
        \label{fig:reward_comparison}
    \end{subfigure}
    \caption{Training dynamics of the proposed rule-compliant VLA method. (a) SFT loss trajectory. (b) Reward curves for the subsequent RLFT stage.}
    \label{fig:training_dynamics}
\end{figure}

\section{RESULTS \& DISCUSSION} \label{sec:results}
This section evaluates the proposed rule-compliant VLA method from both optimization and deployment perspectives. During RL, training dynamics, including reward convergence and rule-learning behaviour are analyzed. The resulting policy is then evaluated on unseen driving scenarios with respect to driving performance, learned driving behaviour, UK Highway Code compliance at both the aggregate and condition-specific levels, and the contribution of individual symbolic rules through ablation.

\vspace{-3mm}
\subsection{Training Performance \& Policy Convergence}
Figure \ref{fig:training_dynamics} summarizes the optimization behaviour of the proposed approach  throughout both SFT and RLFT. During the SFT stage, the training loss shown in Figure \ref{fig:sft_loss_curve} decreases monotonically before converging to a stable plateau, indicating that the model successfully learns the joint distribution of driving actions and their corresponding chain-of-thought reasoning from expert demonstrations.

In the RL stage, Figure \ref{fig:reward_comparison} shows the reward curves for GRPO, GDPO, and Dr.~GRPO. All three methods converge to a stable maximum reward, demonstrating a stable policy optimization. Dr.~GRPO consistently achieves higher reward values compared to GRPO and GDPO. This improvement can be attributed to the removal of sequence-length and standard-deviation normalization terms, which reduces variance introduced by dynamic scaling during optimization. In addition to that, omitting the KL-divergence regularization term appears to yield a smoother reward trajectory, as it reduces competing constraints on policy updates, allowing for more consistent reward maximization over training.

\begin{figure*}[t]
    \centering
    \includegraphics[width=\textwidth]{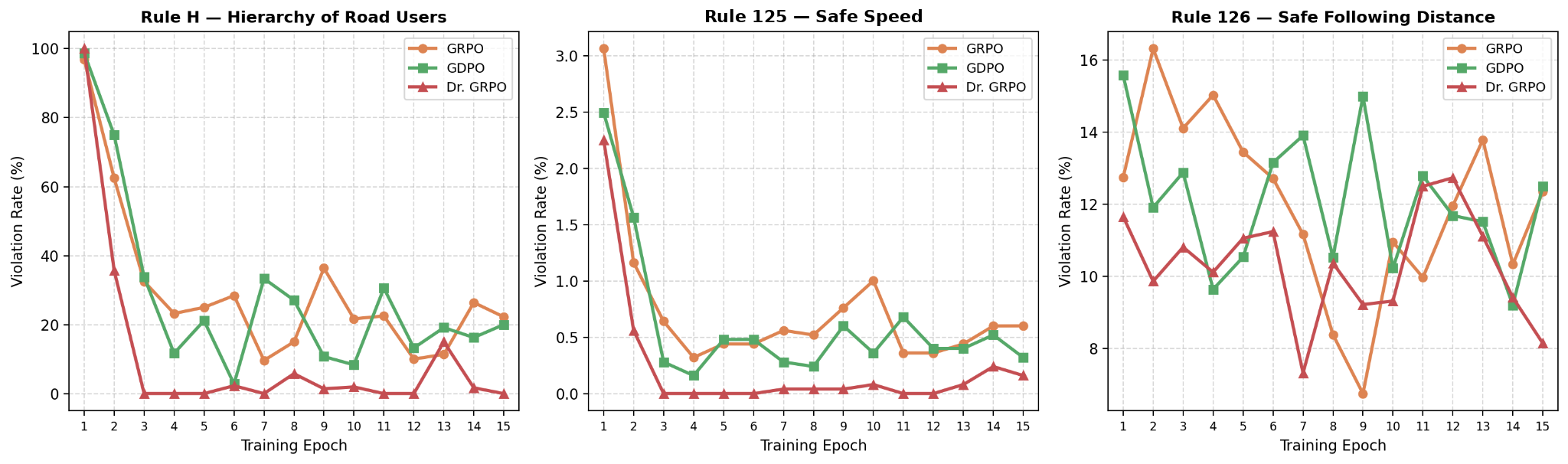}
    \caption{Per-rule violation rates over 15 RLFT epochs. Rates are computed over all symbolic rollout evaluations where the rule was active.}
    \label{fig:rule_violation_rates}
\end{figure*}

Figure \ref{fig:rule_violation_rates} presents the violation rates of each UK Highway Code rule during RLFT, where optimization is initialized from the SFT-pretrained policy. These plots provide insight into how effectively each RLFT approach acquires and retains rule-compliant behaviours during optimization. The three rules show distinct learning patterns.

For \textit{Rule H - Hierarchy of Road Users}, the SFT policy initially violates the constraint nearly 100\% of the time. Here, a violation occurs whenever the vehicle fails to decelerate in a yield scenario, regardless of whether a collision happens. Successfully obeying this rule requires the policy to recognize contextual cues and transition from nominal driving to precautionary slowing. This high failure rate indicates that such context-dependent behavioural adaptation is not reliably acquired through SFT alone. All three RLFT methods reduce violations over the first three epochs. However, their subsequent performance diverges sharply. GRPO and GDPO plateau between 10–35\% with frequent spikes, indicating that learned compliance is periodically lost during optimization. In contrast, Dr.~GRPO reduces violations to almost zero by the third epoch and maintains near-perfect compliance for the remainder of RLFT. 

Under \textit{Rule 125 -- Safe Speed}, the SFT policy shows a low initial violation rate of just 2–3\%. This suggests that basic speed regulation is easily learned from expert demonstrations. Because maintaining speed is a low-level control task rather than a context-dependent yield decision, the policy masters this behaviour before RLFT. The RLFT primarily serves to refine an already competent policy here rather than introducing an entirely new behaviour. All three optimization methods converge rapidly within the first four epochs. However, Dr.~GRPO consistently achieves the lowest residual violation rate, remaining below 0.1\% throughout most of RLFT.

Considering \textit{Rule 126 -- Safe Following Distance}, it remains challenging for all optimization methods. Unlike rules governed by static limits or discrete yielding triggers, Rule 126 demands continuous spacing control that depends not only on the ego vehicle's control, but also on external dynamic factors such as the speed of lead vehicles, traffic light state transitions, and environmental road conditions. Because the required separation margin fluctuates constantly in response to these surrounding changes, the policy must maintain precise, ongoing longitudinal control.

This increased environmental coupling is directly reflected in the training dynamics: all three algorithms exhibit persistent performance oscillations (between ~8\% and 16\%) without a clear convergence trend. Although Dr. GRPO achieves slightly lower violation rates in later epochs, its gains are modest compared to the substantial improvements seen for Rules H and 125. These findings suggest that maintaining a safe following distance requires modeling complex, long-horizon multi-agent interactions that remain difficult for the policy to internalize fully through RLFT alone.

\subsection{Evaluation of Driving Performance}
Table \ref{tab:vla_eval_metrics} compares the action prediction accuracy and rule compliance of the SFT model against the three RLFT methods. To assess stability, each method was evaluated across three independent random seeds controlling dataset ordering and action noise sampling, with results reported as mean $\pm$ standard deviation. Notably, all RLFT variants consistently outperform the SFT policy across every metric. The simultaneous increase in rule compliance alongside reductions in throttle and steering errors demonstrates that optimizing the proposed rule-guided reward does not induce a trade-off between task performance and safety constraint adherence. This is expected because the reward formulation not only encourages compliance with symbolic traffic rules but also preserves an imitation-based objective that discourages unnecessary deviations from the demonstrated driving behaviour. Among the RLFT methods, Dr.~GRPO achieves the best overall performance, obtaining the lowest action prediction errors and the highest pass rate (97.59\%). Although the differences between the RLFT methods are relatively small, the consistent improvements suggest that Dr.~GRPO is able to utilize the reward signal more effectively. One possible explanation is that removing the KL-divergence regularization provides greater flexibility for optimizing rule compliance, while the imitation component of the reward continues to keep the policy close to expert driving behaviour. 

To complement the aggregate pass rates, Table \ref{tab:vla_eval_rules} reports the symbolic compliance score of Equation \ref{eq:rule_score_case}. This metric measures how consistently the policy satisfies each rule when the corresponding traffic scenario is active, where $+1.0$ indicates complete compliance and $-1.0$ denotes complete violation. By restricting the average to frames in which a rule is triggered, the score isolates policy behaviour under the precise driving conditions that each constraint governs. For safe speed rule, the SFT policy already achieves a high baseline score of $0.92$, confirming that maintaining safe speeds is largely captured during imitation learning. RLFT refines this capability further, with all variants reaching perfect compliance $+1.0$. This demonstrates that explicit rule-guided rewards effectively eliminate the remaining edge cases where the SFT policy fails. In comparison, the safe distance rule remains considerably challenging. Although RLFT methods consistently improve compliance from $0.75$ to approximately $0.80$, none of the optimization methods achieves perfect compliance. These observations are consistent with the training dynamics presented previously, where Rule 126 exhibited the slowest convergence and the largest fluctuations throughout RLFT. Maintaining a safe following distance requires continuous reasoning over relative vehicle motion and environmental conditions rather than a single discrete behavioural adjustment, making precise compliance inherently more difficult to learn. The largest improvement is seen for the hierarchy of road users rule. The SFT policy consistently violates this rule and results in the minimum possible score of $-1.0$. This demonstrates that SFT alone fails to reliably initiate the required deceleration behaviour when a must-give-way condition is active. RLFT fundamentally changes this behaviour. Both GRPO and GDPO increase the compliance score to approximately $0.88$, while Dr.~GRPO achieves perfect compliance. This indicates that explicit symbolic supervision is highly effective for learning context-dependent behavioural responses that are rarely acquired reliably through SFT alone.

\begin{figure*}[t]
    \centering
    \includegraphics[width=\textwidth]{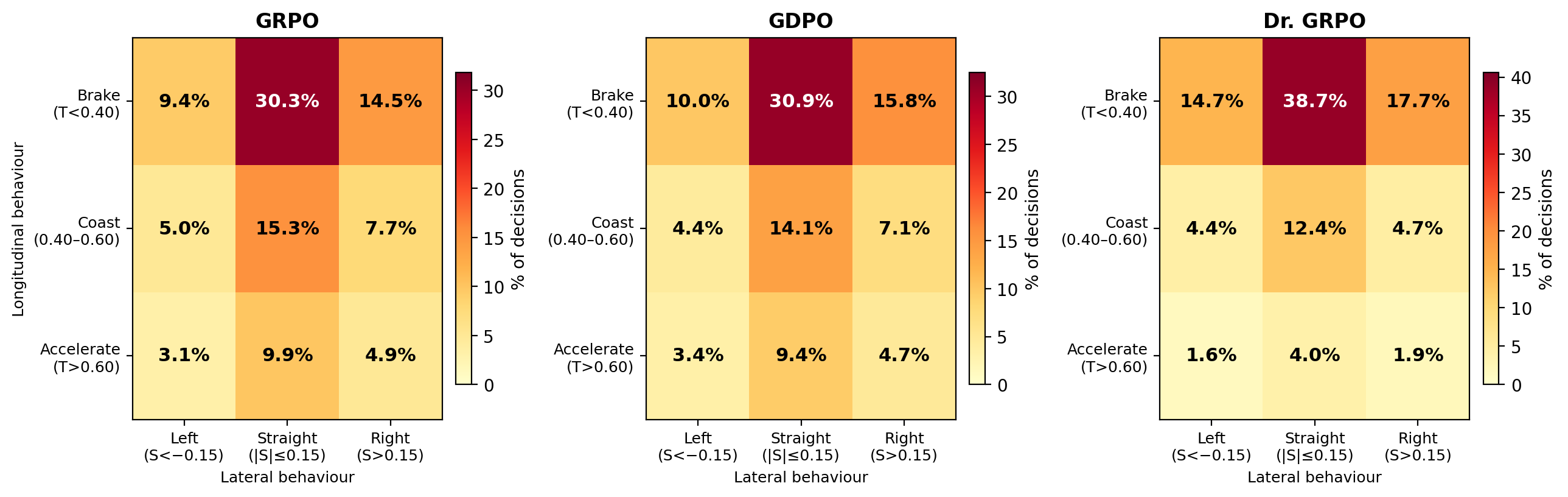}
    \caption{Distribution of longitudinal and lateral driving decisions learned by each RL optimization method. Longitudinal actions are grouped into braking, coasting, and acceleration. Lateral actions are grouped into left, straight, and right steering.}
    \label{fig:driving_behaviour_heatmap}
\end{figure*}

Figure~\ref{fig:driving_behaviour_heatmap} summarizes the distribution of longitudinal (brake, coast, accelerate) and lateral (left, straight, right) control decisions produced by each RLFT method over the evaluation dataset. Among the evaluated methods, Dr.~GRPO exhibits the most noticeable shift in longitudinal behaviour. Compared with GRPO and GDPO, Dr.~GRPO adopts a more conservative longitudinal driving strategy. It performs braking actions more frequently while reducing acceleration across all steering directions. This difference is most apparent during straight driving, where the proportion of braking actions increases from approximately 30\% for GRPO and GDPO to 38.7\% for Dr.~GRPO. At the same time, straight-line acceleration decreases from around 10\% to only 4\%. Similar reductions in acceleration are also observed during left and right steering manoeuvres. These behavioural changes are consistent with the improvements in UK Highway Code compliance reported previously. The encoded traffic rules primarily require the vehicle to reduce speed or initiate timely deceleration when approaching hazards, vulnerable road users, or preceding vehicles. By favouring braking over unnecessary acceleration, Dr.~GRPO is better able to satisfy these safety requirements.

\vspace{-3mm}
\subsection{Rule Compliance Analysis}
\begin{figure*}[t]
    \centering
    \includegraphics[width=\textwidth]{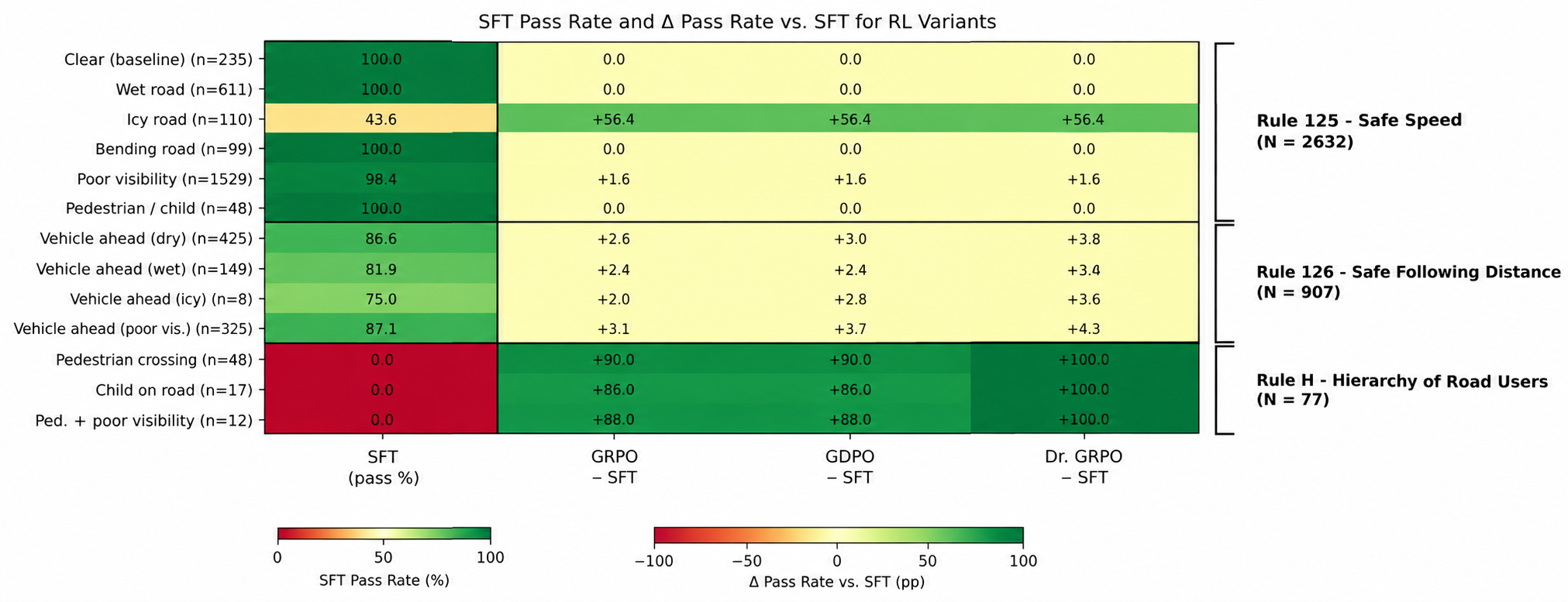}
    \caption{Per-condition rule compliance for the SFT checkpoint and the change ($\Delta$) introduced by each RL variant (GRPO, GDPO, Dr.~GRPO), grouped by UK Highway Code rule. The leftmost column reports the absolute SFT pass rate (\%); the remaining columns report the pass-rate change relative to SFT in percentage points.}
    \label{fig:rule_compliance_heatmap}
    \vspace{-3mm}
\end{figure*}

The previous evaluations provide aggregate measures of rule compliance. They do not reveal how the learned policy behaves under individual traffic scenarios. To address this, a scenario-specific analysis is performed by evaluating each UK Highway Code rule under different driving conditions that activate its predicates. Figure~\ref{fig:rule_compliance_heatmap} presents the corresponding pass rates for the SFT policy together with the improvements achieved by RLFT optimization methods. Several driving scenarios are already handled reliably by the SFT policy. For Rule~125, compliance reaches 100\% under clear weather, wet roads, bending roads, and pedestrian-related conditions, while poor-visibility scenarios achieve 98.4\% compliance. These results suggest that conventional speed regulation under common driving conditions is effectively learned through supervised imitation. However, rule compliance drops significantly under icy-road conditions. Although this scenario is governed by the same safe speed rule, the SFT policy achieves only 43.6\% compliance. This reveals a fundamental limitation of VLA policies trained without explicit rule guidance. SFT policy effectively learns to match expert demonstrations however those demonstrations themselves do not always comply with traffic rules. In contrast, the rule-guided RLFT methods eliminate these violations and improve the compliance by 56.4\% points to achieve perfect performance. Considering the safe following distance rule, across all vehicle-following scenarios, the SFT policy achieves moderate compliance ranging from 75\% to 87\%, indicating that SFT policy captures nominal headway maintenance but remains inconsistent across different environmental conditions. Unlike the Safe Speed rule, no individual scenario reaches complete compliance after RLFT, and only modest improvements of approximately 2--4 percentage points are observed. The most significant behavioural deficiency of the SFT policy is observed for Rule~H. Compliance is zero across all pedestrian- and child-related scenarios, demonstrating a systematic inability to initiate the required precautionary deceleration when a must-give-way condition is triggered. RL substantially improves performance in all three scenarios, indicating that explicit symbolic supervision successfully teaches the policy to adapt its behaviour in the presence of vulnerable road users. While GRPO and GDPO recover compliance to approximately 86--90\%, Dr.~GRPO achieves complete compliance across every evaluated condition, suggesting that the optimized policy generalizes consistently across different must-give-way situations rather than correcting only individual cases.

\vspace{-3mm}
\subsection{Ablation Study}
To assess the individual contribution of each rule to the composite reward signal, an ablation study is conducted with the best-performing RL variant, Dr.~GRPO. The policy is systematically retrained with each UK Highway Code rule omitted from the reward function while maintaining all other training parameters constant. Each ablation variant is trained across three independent random seeds as previously done in Table \ref{tab:vla_eval_metrics}  with performance reported as mean $\pm$ standard deviation.
Table \ref{tab:ablation_rules} demonstrate that each symbolic rule primarily contributes to learning the behaviour it explicitly supervises. Removing Rule~H increases the corresponding violation rate from 0.0\% to 6.1\%, while compliance with Rules~125 and 126 remains largely unaffected. Similarly, excluding Rule~126 produces the largest degradation for the safe following distance rule, increasing its violation rate from 5.9\% to 8.0\%, whereas the remaining rules exhibit only marginal changes. These observations indicate that the proposed reward formulation successfully provides targeted supervision for distinct driving behaviours rather than relying on indirect transfer between rules. The magnitude of the performance degradation varies considerably across rules. Removing Rule~125 results in only a marginal increase in safe speed violations, suggesting that speed regulation is already well captured through the imitation objective and reinforced indirectly by the remaining safety constraints. In contrast, removing either Rule~H or Rule~126 produces substantially larger degradations, indicating that these behaviours depend more strongly on explicit symbolic supervision. This observation agrees with the earlier evaluation, where yielding and following-distance behaviours are identified as the most challenging aspects of policy learning. Removing individual rules produces only minor changes in throttle and steering prediction errors. This suggests that the symbolic reward primarily influences safety-critical decision making rather than the policy's ability to reproduce expert control commands. In other words, the improvements in rule compliance cannot simply be attributed to better action regression but instead arise from learning behaviours that more consistently satisfy the encoded traffic regulations.

\begin{table*}[ht]
\centering
\caption{VLA action regression error and rule pass/violation rates for SFT, GRPO, GDPO, and Dr.~GRPO, reported for about 3,600 held-out samples. To estimate training variance, each method was trained with three independent random seeds controlling dataset ordering and action noise sampling. All runs were initialized from the same supervised fine-tuned checkpoint. Values are reported as mean $\pm$ standard deviation.}
\label{tab:vla_eval_metrics}
\renewcommand{\arraystretch}{1.2}
\begin{tabular}{lcccc}
\hline
\textbf{Metric} & \textbf{SFT} & \textbf{GRPO} & \textbf{GDPO} & \textbf{Dr.\ GRPO} \\
\hline
Throttle MAE (\%)   & $47.38 \pm 0.00$ & $21.19 \pm 0.05$ & $20.77 \pm 0.02$ & $\mathbf{19.16 \pm 0.02}$ \\
Steer MAE (\%)      & $18.70 \pm 0.02$ & $9.17 \pm 0.10$  & $7.11 \pm 0.10$  & $\mathbf{5.36 \pm 0.20}$ \\
Throttle RMSE (\%)  & $48.31 \pm 0.05$ & $23.71 \pm 0.08$ & $23.54 \pm 0.05$ & $\mathbf{23.17 \pm 0.00}$ \\
Steer RMSE (\%)     & $20.96 \pm 0.10$ & $13.74 \pm 0.10$ & $12.95 \pm 0.10$ & $\mathbf{12.74 \pm 0.10}$ \\
Euclidean MAE (\%)  & $73.39 \pm 0.05$ & $43.07 \pm 0.05$ & $42.30 \pm 0.05$ & $\mathbf{40.57 \pm 0.05}$ \\
Euclidean RMSE (\%) & $81.07 \pm 0.05$ & $45.82 \pm 0.20$ & $45.43 \pm 0.20$ & $\mathbf{45.01 \pm 0.10}$ \\
Pass Rate (\%)      & $92.93 \pm 1.00$ & $97.34 \pm 0.50$ & $97.39 \pm 0.50$ & $\mathbf{97.59 \pm 0.50}$ \\
Violate Rate (\%)   & $7.07 \pm 1.00$  & $2.66 \pm 0.50$  & $2.61 \pm 0.50$  & $\mathbf{2.41 \pm 0.50}$ \\
\hline
\end{tabular}
\end{table*}

\begin{table}[h]
\centering
\caption[Per-rule symbolic compliance scores for SFT, GRPO, GDPO and Dr.~GRPO]{%
Symbolic compliance score (Eq.~\eqref{eq:rule_score_case}) for each UK Highway Code rule, 
averaged over frames in which the rule is active, for SFT, GRPO, GDPO, and Dr.~GRPO. 
Higher is better; $+1.0$ indicates satisfaction in every active frame.}
\label{tab:vla_eval_rules}
\renewcommand{\arraystretch}{1.2}
\resizebox{\linewidth}{!}{%
\begin{tabular}{lcccc}
\toprule
\textbf{Rule} & \textbf{SFT} & \textbf{GRPO} & \textbf{GDPO} & \textbf{Dr.\ GRPO} \\
\midrule
Safe Speed Rule Score             & 0.9200  & \textbf{1.0000} & \textbf{1.0000} & \textbf{1.0000} \\
Safe Distance Rule Score          & 0.7468  & 0.7848 & \textbf{0.7975}  & \textbf{0.7975} \\
Hierarchy of Road Users Rule Score & -1.0000 & \textbf{0.8820} & \textbf{0.8820} & \textbf{1.0000} \\
\toprule
\end{tabular}%
}
\end{table}

\begin{table*}[t]
\centering
\caption{Rule ablation study for Dr.~GRPO. Each row corresponds to a variant trained with a different subset of rules in the reward signal. Violation rates and control accuracy (MAE) are reported on the held-out test set. Values are reported as mean $\pm$ standard deviation over three random seeds. $\downarrow$ denotes lower is better.}
\label{tab:ablation_rules}
\renewcommand{\arraystretch}{1.2}
\begin{tabular}{lccccc}
\toprule
& \multicolumn{2}{c}{\textbf{Control Accuracy}} & \multicolumn{3}{c}{\textbf{Rule Violation Rate $\downarrow$}} \\
\cmidrule(lr){2-3} \cmidrule(lr){4-6}
\textbf{Rules in Reward} &
\textbf{Throttle MAE (\%) $\downarrow$} &
\textbf{Steer MAE (\%) $\downarrow$} &
\textbf{Rule H} &
\textbf{Rule 125} &
\textbf{Rule 126} \\
\midrule
Rule-H + Rule-125 + Rule-126 &
$\mathbf{19.29 \pm 0.02}$ &
$6.50 \pm 0.02$ &
$\mathbf{0.00 \pm 0.10}\%$ &
$\mathbf{0.10 \pm 0.10}\%$ &
$5.90 \pm 0.30\%$ \\

Rule-125 + Rule-126 &
$21.04 \pm 0.02$ &
$\mathbf{5.67 \pm 0.01}$ &
$6.10 \pm 0.30\%$ &
$0.20 \pm 0.10\%$ &
$\mathbf{5.50 \pm 0.20}\%$ \\

Rule-H + Rule-126 &
$21.54 \pm 0.02$ &
$6.75 \pm 0.02$ &
$5.40 \pm 0.20\%$ &
$0.20 \pm 0.10\%$ &
$5.70 \pm 0.20\%$ \\

Rule-H + Rule-125 &
$20.61 \pm 0.02$ &
$6.98 \pm 0.02$ &
$4.40 \pm 0.20\%$ &
$\mathbf{0.10 \pm 0.10}\%$ &
$8.00 \pm 0.40\%$ \\
\bottomrule
\end{tabular}
\end{table*}

\section{Conclusion}
\label{sec:conclusion}

This paper presented a rule-guided VLA framework for autonomous driving that combines supervised learning from expert demonstrations with symbolic traffic-rule feedback during RL fine-tuning. Selected UK Highway Code provisions are formalised as executable FOL rules using PyReason to evaluate the projected consequences of candidate control actions. The resulting reward encourages compliance with the encoded rules while penalising deviations from expert actions. Offline experiments on held-out CARLA data show that the three evaluated RL methods---GRPO, GDPO, and Dr.~GRPO---improve both action prediction accuracy and encoded-rule compliance over the SFT baseline. Dr.~GRPO achieves the highest overall rule pass rate, increasing it from 92.9\% for the SFT baseline to 97.6\%, with zero observed violations of the encoded road-user-priority check in the evaluated samples. These findings support explicit symbolic feedback as a complement to demonstration-based learning, allowing traffic-rule criteria to directly influence VLA policy refinement.

Several limitations remain. The evaluation is conducted entirely with pre-collected multimodal CARLA data, and the degree to which these findings transfer to real-world driving conditions remains an open question. The symbolic rule set is also limited to three checks covering selected UK Highway Code provisions, and broader coverage of traffic regulations, as well as adaptation to other regulatory frameworks, warrants future investigation. Future work will focus on extending the symbolic rule set to cover a wider range of traffic scenarios, exploring closed-loop evaluation in CARLA to assess the impact of rule-guided actions on longer driving trajectories, and investigating the transferability of the framework to real-world driving datasets and regulatory frameworks beyond the UK Highway Code.
\section*{ACKNOWLEDGMENT}
The first author thanks research funding support from UK Engineering and Physical Science Research Council. 

\section{Biography Section}
\begin{IEEEbiography}[{\includegraphics[width=1in,height=1.25in,clip,keepaspectratio]{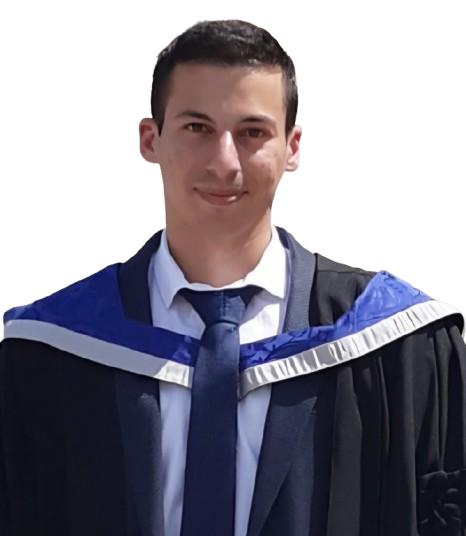}}]{Barkin Dagda}
received the B.Eng. degree in Mechanical Engineering and the M.Sc. degree in Electronic Engineering from the University of Surrey, Guildford, U.K., in 2021 and 2022, respectively. He is currently pursuing the Ph.D. degree in Automotive Engineering at the University of Surrey, within the Connected and Autonomous Vehicles (CAV) Lab. His PhD  focuses on vision-based foundation models for autonomous vehicles. With experience in academia and industry, he has worked on AI-driven robotics, navigation, and automation
\end{IEEEbiography}
\vspace{-10mm}
\begin{IEEEbiography}[{\includegraphics[width=1in,height=1.25in,clip,keepaspectratio]{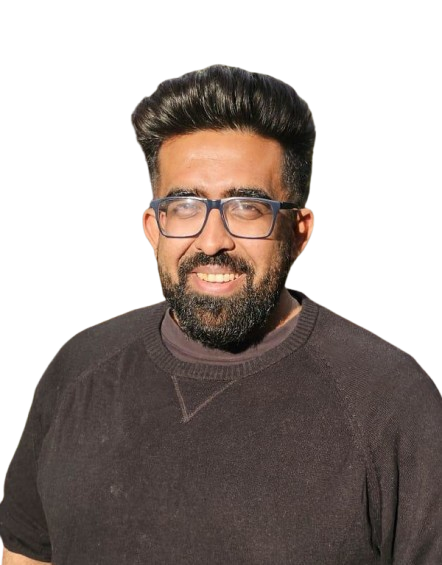}}]{Vinal Asodia}
is a Ph.D. Researcher in the Connected and Autonomous Vehicles (CAV) Lab at the University of Surrey, Guildford, U.K. He received the B.Sc. degree in Computer Science from the University of Surrey. His research focuses on developing end-to-end deep reinforcement learning methods for autonomous driving systems that are explainable and aligned with human values. His research interests include enhancing the transparency, safety, and social acceptability of AI-driven autonomous systems, contributing to the reliability of these technologies in real-world applications. He has published his work in the IEEE International Conference on Robotics and Automation (ICRA).
\end{IEEEbiography}
\vspace{-10mm}
\begin{IEEEbiography}
[{\includegraphics[width=1in,height=1.25in,clip,keepaspectratio]{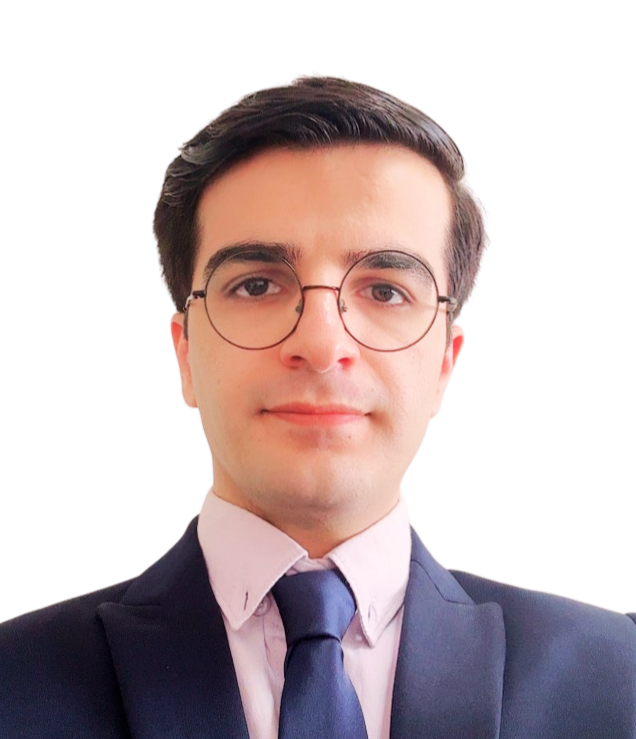}}]{Iman Sharifi} is a Ph.D. Candidate in Mechanical and Aerospace Engineering at George Washington University. He
received his M.Sc. in Mechanical Engineering from Sharif University of Technology in 2022. His research focuses on neuro-symbolic artificial intelligence, multi-agent reinforcement learning, and large language and vision models in autonomous driving and aviation applications. He is currently a visiting researcher at Stanford Intelligent Systems Laboratory (SISL), Department of Aeronautics and Astronautics, Stanford University. He has conducted research funded by NASA’s University Leadership Initiative (ULI). 
\end{IEEEbiography}
\vspace{-10mm}
\begin{IEEEbiography}[{\includegraphics[width=1in,height=1.25in,clip,keepaspectratio]{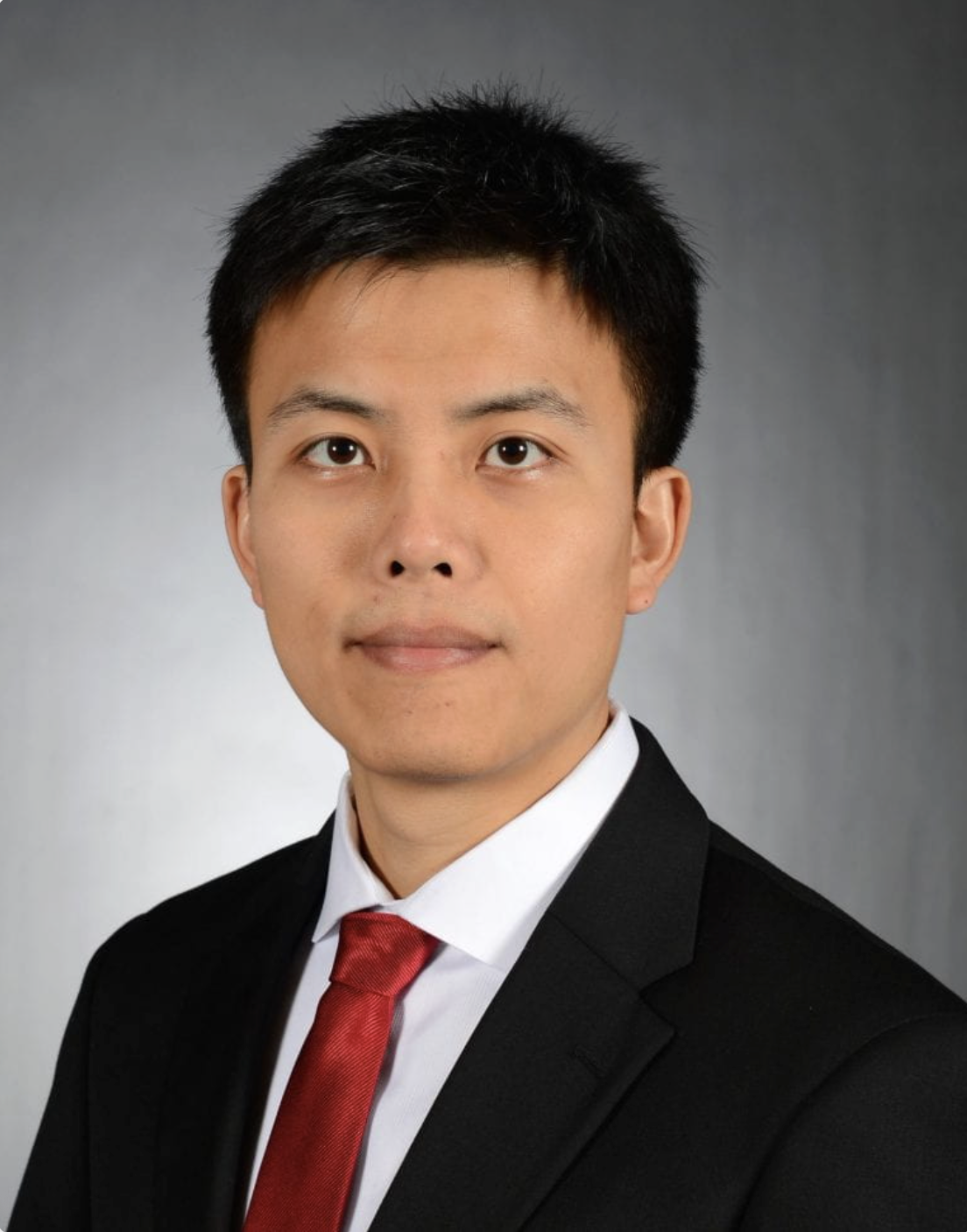}}]{Peng Wei}
received his PhD degree in aerospace engineering from Purdue University in 2013. He is currently a Professor at the Department of Mechanical and Aerospace Engineering, George Washington University, with courtesy appointments at the Electrical and Computer Engineering Department and the Computer Science Department. His current research interests include safety, efficiency and scalability of decision-making systems in complex, uncertain, and dynamic environments.
\end{IEEEbiography}
\vspace{-10mm}
\begin{IEEEbiography}[{\includegraphics[width=1in,height=1.25in,clip,keepaspectratio]{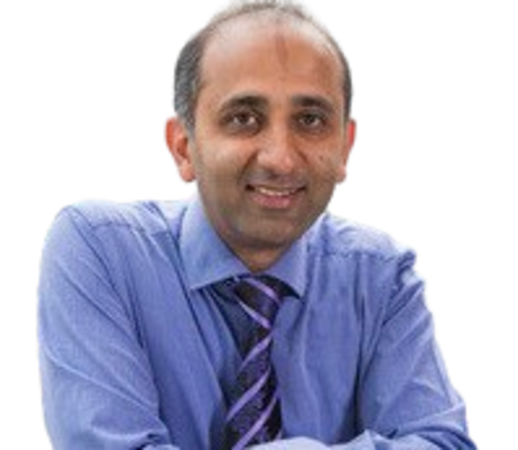}}]{Saber Fallah}
is a Professor of Safe AI and Autonomy at the University of Surrey, UK, where he directs the Connected and Autonomous Vehicles Lab (CAV-Lab). His research centres on safe and trustworthy artificial intelligence for autonomous systems, particularly autonomous vehicles and robotics. His work combines reinforcement learning, neuro-symbolic AI, vision-language-action models, and safety assurance to develop autonomous systems capable of reliable reasoning and decision-making in complex environments. He has led research funded by EPSRC, Innovate UK, the European Commission, and industry, and works extensively across academia, industry, and policy on the development and deployment of trustworthy autonomous technologies.
\end{IEEEbiography}

\begin{thebibliography}{50}

\bibliographystyle{IEEEtran}

\bibitem{hu2026vla}
T. Hu, X. Liu, S. Wang, Y. Zhu, A. Liang, L. Kong, G. Zhao, Z. Gong, J. Cen, Z. Huang, X. Hao, L. Li, H. Song, X. Li, J. Ma, S. Shen, J. Zhu, D. Tao, Z. Liu, and J. Liang,
``Vision-Language-Action Models for Autonomous Driving: Past, Present, and Future,''
arXiv preprint arXiv:2512.16760, 2026.
[Online]. Available: https://arxiv.org/abs/2512.16760


\bibitem{chen2024e2e}
L. Chen, P. Wu, K. Chitta, B. Jaeger, A. Geiger, and H. Li,
``End-to-End Autonomous Driving: Challenges and Frontiers,''
\emph{IEEE Transactions on Pattern Analysis and Machine Intelligence},
vol. 46, no. 12, pp. 10164--10183, Dec. 2024,
doi: 10.1109/TPAMI.2024.3435937.


\bibitem{jiang2025}
S.~Jiang, Y.~Zhang, Y.~Wang, Z.~Li, C.~Sima, J.~F.~P.~Kooij, G.~P.~R.~Papandreou, H.~Zhao, and L.~Chen,
``A Survey on Vision-Language-Action Models for Autonomous Driving,''
in \emph{2025 IEEE/CVF International Conference on Computer Vision Workshops (ICCVW)}, 2025, pp. 4583--4595, doi: 10.1109/ICCVW69036.2025.00476.

\bibitem{yildirim2025}
M.~Yildirim, B. Dagda, V. Asodia and S. Fallah, ``Highwayllm: Decision-making and navigation in highway driving with rl-informed language model,'' \emph{Robotics and Autonomous Systems}, 2025, 193, p.105114.


\bibitem{drivelm}
S. Shao, Z. Zou, L. Chen, Y. Li, J. Yan, B. Jaeger, K. Chitta, P. Wu, Y. Zhang, and H. Li,
``DriveLM: Driving with Graph Visual Question Answering,''
in \emph{Proc. IEEE/CVF Conf. Computer Vision and Pattern Recognition (CVPR)}, 2024.
[Online]. Available: https://arxiv.org/abs/2312.14150


\bibitem{lmdrive}
H. Wang, C. Sima, J. Dai, H. Hu, L. Chen, S. Wu, H. Zhang, and J. Yan,
``LMDrive: Closed-Loop End-to-End Driving with Large Language Models,''
in \emph{Proc. Conf. Robot Learning (CoRL)}, 2023.
[Online]. Available: https://arxiv.org/abs/2312.07442


\bibitem{drivegpt4}
Z. Xu, Y. Zhang, E. Xie, Z. Zhao, Y. Guo, K.-Y. K. Wong, Z. Li, and H. Zhao,
``DriveGPT4: Interpretable End-to-End Autonomous Driving Via Large Language Model,''
\emph{IEEE Robotics and Automation Letters}, vol. 9, no. 10, pp. 8186--8193, Oct. 2024,
doi: 10.1109/LRA.2024.3440097.


\bibitem{guo2024drivemllm}
X.~Guo, R.~Zhang, Y.~Duan, Y.~He, C.~Zhang, S.~Liu, and L.~Chen,
``DriveMLLM: A Benchmark for Spatial Understanding with Multimodal Large Language Models in Autonomous Driving,''
arXiv preprint arXiv:2411.13112, 2024.


\bibitem{jiang2024senna}
B.~Jiang, S.~Chen, B.~Liao, X.~Zhang, W.~Yin, Q.~Zhang, C.~Huang, W.~Liu, and X.~Wang,
``Senna: Bridging Large Vision-Language Models and End-to-End Autonomous Driving,''
arXiv preprint arXiv:2410.22313, 2024.


\bibitem{xie2025drivebench}
S.~Xie, L.~Kong, Y.~Dong, C.~Sima, W.~Zhang, Q.~A.~Chen, Z.~Liu, and L.~Pan,
``Are VLMs Ready for Autonomous Driving? An Empirical Study from the Reliability, Data, and Metric Perspectives,''
in \emph{Proceedings of the IEEE/CVF International Conference on Computer Vision (ICCV)}, pp. 6585--6597, 2025.


\bibitem{zhou2025autovla}
Z.~Zhou, T.~Cai, S.~Z.~Zhao, Y.~Zhang, Z.~Huang, B.~Zhou, and J.~Ma,
``AutoVLA: A Vision-Language-Action Model for End-to-End Autonomous Driving with Adaptive Reasoning and Reinforcement Fine-Tuning,''
in \emph{Advances in Neural Information Processing Systems (NeurIPS)}, vol. 39, 2025.


\bibitem{zhou2025opendrivevla}
X. Zhou, X. Han, F. Yang, Y. Ma, and A. C. Knoll,
``OpenDriveVLA: Towards End-to-End Autonomous Driving with Large Vision Language Action Model,''
\emph{Proceedings of the AAAI Conference on Artificial Intelligence}, vol. 40, no. 16, pp. 13782--13790, 2026,
doi: 10.1609/aaai.v40i16.38386.


\bibitem{dvlm-ad}
Y.~Ma, Y.~Cao, W.~Ding, S.~Zhang, Y.~Wang, B.~Ivanovic, M.~Jiang, M.~Pavone, and C.~Xiao,
``dVLM-AD: Enhance Diffusion Vision-Language-Model for Driving via Controllable Reasoning,''
arXiv preprint arXiv:2512.04459, 2025.
[Online]. Available: https://arxiv.org/abs/2512.04459


\bibitem{simlingo}
K. Renz, L. Chen, E. Arani, and O. Sinavski,
``SimLingo: Vision-Only Closed-Loop Autonomous Driving with Language-Action Alignment,''
in \emph{Proc. IEEE/CVF Conf. Computer Vision and Pattern Recognition (CVPR)}, 2025, pp. 11993--12003,
doi: 10.1109/CVPR52734.2025.01120.


\bibitem{cfvla}
F.~Du, F.~Yan, J.~Wu, X.~Xu, W.~Zhang, W.~Wang, Y.~Guo, B.~Qian, Z.~He, F.~Wang, and H.~Yang,
``CF-VLA: Efficient Coarse-to-Fine Action Generation for Vision-Language-Action Policies,''
arXiv preprint arXiv:2604.24622, 2026.
[Online]. Available: https://arxiv.org/abs/2604.24622


\bibitem{perceptwam}
J. Han, M. Tian, J. Zhu, F. He, H. Zhang, S. Guo, D. Zhu, H. Tang, P. Xu, Y. Guo, M. Niu, H. Zhu, Q. Dong, X. Yan, S. Dong, L. Hou, Q. Huang, X. Jia, and H. Xu,
``Percept-WAM: Perception-Enhanced World-Awareness-Action Model for Robust End-to-End Autonomous Driving,''
in \emph{Proc. IEEE/CVF Conf. Computer Vision and Pattern Recognition (CVPR)}, 2026.


\bibitem{autodrive}
Z.~Yuan, C.~Qian, J.~Tang, R.~Chen, Z.~Song, L.~Sun, X.~Chu, Y.~Cai, D.~Zhang, and S.~Li,
``AutoDrive-R$^2$: Incentivizing Reasoning and Self-Reflection Capacity for VLA Model in Autonomous Driving,''
arXiv preprint arXiv:2509.01944, 2026.
[Online]. Available: https://arxiv.org/abs/2509.01944


\bibitem{Sharifi_2}
I.~Sharifi, M.~Yildirim, and S.~Fallah, ``Toward Safe Autonomous Highway Driving Policies using a Neuro-Symbolic Deep Reinforcement Learning Approach,'' \emph{Transportation Research Record: Journal of the Transportation Research Board}, vol. 2680, no.~1, pp. 416--431, Sep. 2025.


\bibitem{Sharifi_1}
I.~Sharifi, M.~Yildirim, and S.~Fallah, ``Symbolic Imitation Learning: From Black-Box to Explainable Driving Policies,'' \emph{Applied Sciences}, vol.~15, no.~23, p. 12464, Nov. 2025.


\bibitem{asodia2026}
V.~Asodia, I.~Sharifi, and S.~Fallah,
``Reinforcement Learning Enhancement Using Vector Semantic Representation and Symbolic Reasoning for Human-Centered Autonomous Emergency Braking,''
arXiv preprint arXiv:2602.05079, 2026.
[Online]. Available: https://arxiv.org/abs/2602.05079


\bibitem{safeauto}
J. Zhang, X. Yang, T. Wang, Y. Yao, A. Petiushko, and B. Li,
``SafeAuto: Knowledge-Enhanced Safe Autonomous Driving with Multimodal Foundation Models,''
in \emph{Proceedings of the 42nd International Conference on Machine Learning (ICML)}, PMLR, 2025, pp. 76497--76517.


\bibitem{pyreason}
D. Aditya, K. Mukherji, S. Balasubramanian, A. Chaudhary, and P. Shakarian,
``PyReason: Software for Open World Temporal Logic,''
\emph{CEUR Workshop Proceedings}, vol. 3433, 2023.


\bibitem{ukhighwaycode}
UK Department for Transport,
\emph{The Highway Code},
GOV.UK, 2015, updated Oct. 2025.
[Online]. Available: https://www.gov.uk/guidance/the-highway-code


\bibitem{deepseek}
D. Guo, D. Yang, H. Zhang, L. Song, R. Zhang, R. Xu, Q. Zhu, S. Ma, P. Wang, Z. Shao, J. Yuan, Y. K. Li, W. Gao, S. Wu, Y. Wu, and D. Guo,
``DeepSeek-R1 incentivizes reasoning in LLMs through reinforcement learning,''
\emph{Nature}, vol. 645, no. 8081, pp. 633--638, Sept. 2025,
doi: 10.1038/s41586-025-09422-z.


\bibitem{grpo}
Z.~Shao, P.~Wang, Q.~Zhu, R.~Xu, J.~Song, X.~Bi, H.~Zhang, M.~Zhang, Y.~K.~Li, Y.~Wu, and D.~Guo,
``DeepSeekMath: Pushing the Limits of Mathematical Reasoning in Open Language Models,''
arXiv preprint arXiv:2402.03300, 2024.
[Online]. Available: https://arxiv.org/abs/2402.03300


\bibitem{gdpo}
S.-Y. Liu, X. Dong, X. Lu, S. Diao, P. Belcak, M. Liu, M.-H. Chen, H. Yin, Y.-C. F. Wang, K.-T. Cheng, Y. Choi, J. Kautz, and P. Molchanov,
``GDPO: Group reward-Decoupled Normalization Policy Optimization for Multi-reward RL Optimization,''
in \emph{Proceedings of the 43rd International Conference on Machine Learning (ICML)}, 2026.


\bibitem{drgrpo}
Z.~Liu, C.~Chen, W.~Li, P.~Qi, T.~Pang, C.~Du, W.~S. Lee, and M.~Lin,
``Understanding r1-zero-like training: A critical perspective,''
in \emph{Conference on Language Modeling (COLM)}, 2025.


\bibitem{russelstuart}
S.~J.~Russell and P.~Norvig, \emph{Artificial Intelligence: A Modern Approach}, 4th global ed. London, U.K.: Pearson, 2022.


\bibitem{luo2026}
Y.~Luo, Q.~Chen, F.~Li, S.~Xu, J.~Liu, Z.~Song, Z.-x. Yang, and F.~Wen,
``Unleashing VLA Potentials in Autonomous Driving via Explicit Learning from Failures,''
arXiv preprint arXiv:2603.01063, 2026.
[Online]. Available: https://arxiv.org/abs/2603.01063


\bibitem{carla}
A.~Dosovitskiy, G.~Ros, F.~Codevilla, A.~Lopez, and V.~Koltun,
``CARLA: An Open Urban Driving Simulator,''
in \emph{Proceedings of the 1st Annual Conference on Robot Learning (CoRL)}, 2017, pp. 1--16.

\bibitem{astrom2021feedback}
K.~J. {\AA}str{\"o}m and R.~M. Murray, \emph{Feedback Systems: An Introduction for Scientists and Engineers}, 2nd~ed.\hskip 1em plus 0.5em minus 0.4em\relax Princeton, NJ: Princeton University Press, 2021.

\bibitem{kim2024openvla}
M. J. Kim, K. Pertsch, S. Karamcheti, T. Xiao, A. Balakrishna, S. Nair, R. Rafailov, E. Foster, G. Lam, P. Sanketi, Q. Vuong, T. Kollar, B. Burchfiel, R. Tedrake, D. Sadigh, S. Levine, P. Liang, and C. Finn,
``OpenVLA: An Open-Source Vision-Language-Action Model,''
in \emph{Proceedings of the 8th Conference on Robot Learning (CoRL)}, 2024.


\bibitem{rajamani2012vehicle}
R.~Rajamani,
\emph{Vehicle Dynamics and Control},
2nd~ed.,
Mechanical Engineering Series.
New York, NY, USA: Springer, 2012,
doi: 10.1007/978-1-4614-1433-9.

\end{thebibliography}
\end{document}